\documentclass[letterpaper]{article} 
\usepackage{aaai2026}  
\usepackage{times}  
\usepackage{helvet}  
\usepackage{courier}  
\usepackage[hyphens]{url}  
\usepackage{graphicx} 
\usepackage{natbib}  
\usepackage{caption} 
\usepackage{algorithm}
\usepackage{algorithmic}

\usepackage{newfloat}
\usepackage{listings}
\DeclareCaptionStyle{ruled}{labelfont=normalfont,labelsep=colon,strut=off} 
\floatstyle{ruled}
\newfloat{listing}{tb}{lst}{}
\floatname{listing}{Listing}
\usepackage{amsmath}
\usepackage{amssymb}
\usepackage{booktabs}       
\usepackage{multirow}

\title{Semi-Supervised High Dynamic Range Image Reconstructing via\\ Bi-Level Uncertain Area Masking}
\author{
    Wei Jiang\equalcontrib,
    Jiahao Cui\equalcontrib,
    Yizheng Wu,
    Zhan Peng,
    Zhiyu Pan\thanks{Corresponding author.},
    Zhiguo Cao
}
\affiliations{
    School of AIA, Huazhong University of Science and Technology\\

    \{wei\_jiang,smartcjh,yzwu21,peng\_zhan,zhiyupan,zgcao\}@hust.edu.cn
}

\usepackage{bibentry}

\def\onedot{.}
\def\eg{\emph{e.g}\onedot}
\def\ie{\emph{i.e}\onedot}

\begin{document}

\maketitle

\begin{abstract}
Reconstructing high dynamic range (HDR) images from low dynamic range (LDR) bursts plays an essential role in the computational photography. Impressive progress has been achieved by learning-based algorithms which require LDR-HDR image pairs. However, these pairs are hard to obtain, which motivates researchers to delve into the problem of annotation-efficient HDR image reconstructing: how to achieve comparable performance with limited HDR ground truths (GTs). This work attempts to address this problem from the view of semi-supervised learning where a teacher model generates pseudo HDR GTs for the LDR samples without GTs and a student model learns from pseudo GTs. Nevertheless, the confirmation bias, \ie, the student may learn from the artifacts in pseudo HDR GTs, presents an impediment. To remove this impediment, an uncertainty-based masking process is proposed to discard unreliable parts of pseudo GTs at both pixel and patch levels, then the trusted areas can be learned from by the student. With this novel masking process, our semi-supervised HDR reconstructing method not only outperforms previous annotation-efficient algorithms, but also achieves comparable performance with up-to-date fully-supervised methods by using only $6.7\%$ HDR GTs. 
\end{abstract}

\begin{links}
    \link{Code}{https://github.com/JW20211/SmartHDR}
\end{links}

\begin{figure}[!t]
\centering
\includegraphics[width=\linewidth]{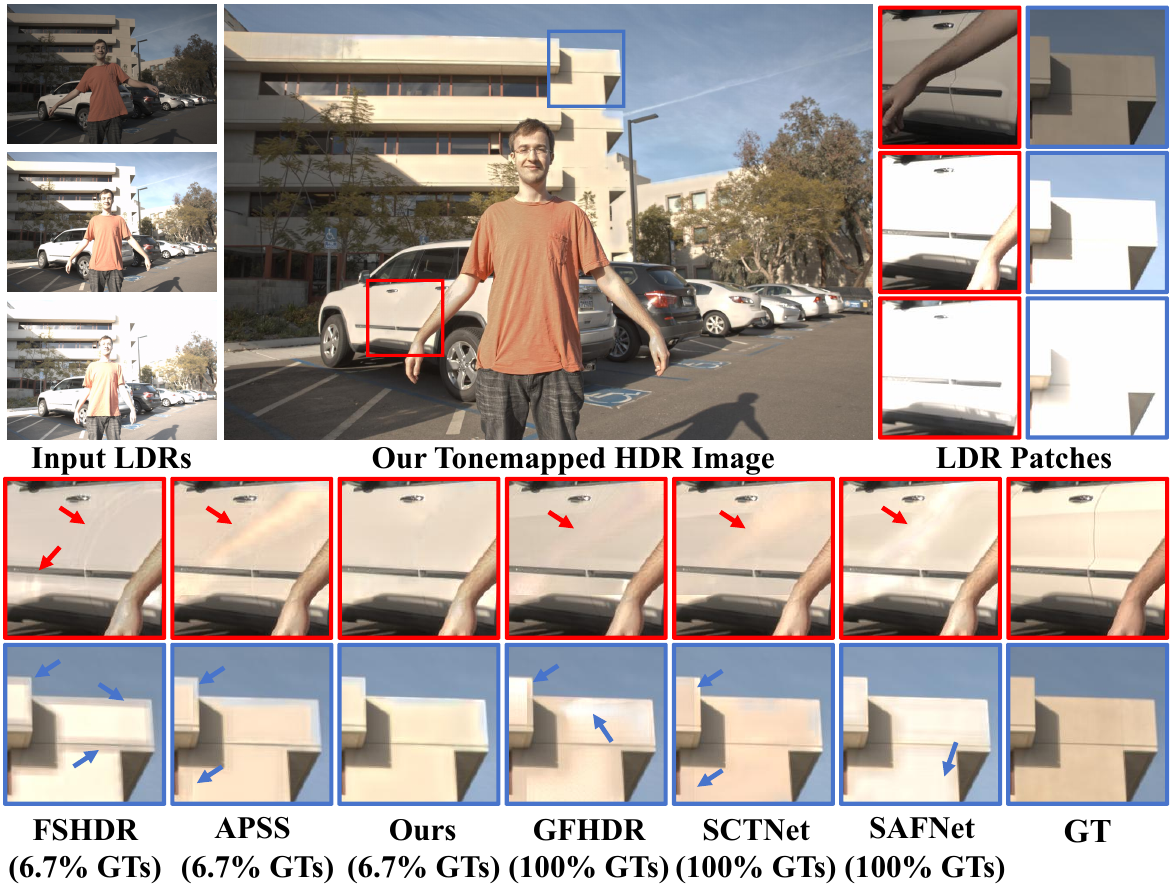}

\caption{\textbf{Qualitative comparison with prior arts.} The illustrated example shows that, trained with only $6.7\%$ GTs, our method outperforms previous annotation-efficient methods and achieve comparable or even better qualitative performance compared with methods trained with all GTs.}
\label{fig_0}

\end{figure}

\section{Introduction}

Human eyes adjust the local exposure gain dynamically with retinal ganglion cells. Hence, views with high dynamic range (HDR) illumination can be presented in the brain~\cite{larson1998logluv}. Current digital imaging systems, where sensors have limited light sensitivity ranges~\cite{hoefflinger2007high}, can only reflect the absolute light intensity~\cite{niu2021hdr}, which produces low dynamic range (LDR) images. For the goal of \textit{what you see is what you get}, researchers try to fuse the differently exposed LDR images into HDR images~\cite{kalantari2017deep}. 
However this pipeline causes ghosting artifacts for dynamic scenes.

To remove ghosting artifacts, most modern methods fall into a learning-based paradigm~\cite{liu2021adnet,chung2022high,liu2022ghost,yan2023unified,yan2023toward,hu2024generating,kong2024safnet}. These methods require paired training samples: LDR images with corresponding HDR ground truths (GTs). The HDR GTs are usually obtained with the help of expensive specially designed imaging systems~\cite{reinhard2010high}, or by controlling the scene motion~\cite{kalantari2017deep,chen2021hdr}. This makes it challenging to collect training samples with HDR GTs at scale, which curbs the development of this task. To tackle this issue, recent advances delve into the problem of annotation-efficient HDR image reconstructing: \textit{how to train a HDR reconstructing model with limited HDR GTs?} 

The FSHDR~\cite{prabhakar2021labeled} generates pseudo HDR GTs and inverts them into synthesized LDR images to obtain paired training samples. However, the domain gap between the real and synthesized LDR images may confuse the model, which leads to the artifacts shown in the first column of Fig.~\ref{fig_0}. The SMAE~\cite{yan2023smae} follows the teacher-student manner and proposes an adaptive pseudo-label selection strategy (APSS) based on the similarity between the pseudo GTs and the reference images. This strategy ignores the saturation area, which is the key challenge of HDR reconstructing. Besides, the teacher and student in SMAE have a marginal gap. Previous self-/semi-supervised learning methods~\cite{rasmus2015semi,laine2016temporal,tarvainen2017mean,sohn2020fixmatch} have demonstrated that the large enough gap between the teacher and student matters for learning new knowledge from unlabeled data. Therefore, the limitations of prior arts can be summarized as: (a) the sub-optimal choice of the teacher model; and (b) the incomplete pseudo GT selecting strategy.

For the limitation (a), we develop a new framework 
in which the student model is updated by the regular gradient descent, and the teacher model is updated by the exponential moving average (EMA) of student weights. In this manner, the teacher can be seen as a temporal ensembling version of the student.
This operation provides a better teacher that can generate high-quality pseudo HDR GTs~\cite{tarvainen2017mean}. In terms of the limitation (b), we propose to mask the pseudo GTs for training according to their reliability rather than hand-crafted rules. Inspired by 
\cite{he2019bounding}, we add an uncertainty estimation branch to our HDR reconstructing model and append Kullback-Leibler (KL) divergence 
 loss to the original loss. The outputs of the model include not only the HDR images but also uncertainty maps which indicate the reliability of the HDR prediction at pixel level. Based on uncertainty maps, we design a dual threshold policy to mask out the unreliable parts of pseudo GTs 
 at both the patch and pixel levels. This bi-level uncertain area masking policy ensures that reliable and informative pseudo GTs can be learned. 

Experimental results illustrate that our framework learns LDR-to-HDR mapping more efficiently with limited HDR GTs: our method can achieve comparable performance with current state-of-the-art fully-supervised methods with only $6.7\%$ HDR GTs, as shown in Fig.~\ref{fig_0}.
The contributions of our work are three-fold: 

$\bullet$ We establish a semi-supervised HDR image reconstructing framework firstly taking the predicted uncertainty of pseudo GTs into account.
 
$\bullet$ We propose a novel bi-level uncertain area masking policy to make sure that trusted pseudo HDR GTs can be learned, which leads to a better performance.

$\bullet$ Our method achieves new state-of-the-art performance under the annotation-efficient setting on public benchmarks.

\section{Related Works}

\subsection{Multi-Frame HDR Image Reconstructing}
To reconstruct the ghost-free HDR image from a set of unaligned LDR images with varied exposures, traditional methods 
align LDR images via homographies~\cite{tomaszewska2007image} or optical flow~\cite{bogoni2000extending,zimmer2011freehand}; some other works reject~\cite{grosch2006fast,oh2014robust,lee2014ghost} misaligned regions in LDR images before fusing them. However these method leads to 
unsatisfied results when large 
motion and saturation occur.

With the rise of the Deep Neural Networks (DNNs), the DNN-based HDR reconstructing methods~\cite{niu2021hdr,chung2022high,li2022uphdr,liu2022ghost,tel2023alignment,cui2024exposure, hu2024generating, kong2024safnet} achieve a significant performance improvement and better visual results.
The pioneer works propose to explicitly align LDR images~\cite{liu2009beyond, wu2018deep} before fusing by CNN network~\cite{kalantari2017deep,wu2018deep}.
However, explicit alignment may fail due to the fact that the misalignment is difficult to be learned.
To solve this problem, some works adopt spatial attention to replace the explicit alignment and fuse the attention-based features to reconstruct HDR images~\cite{yan2019attention,liu2022ghost, yan2023smae}, others propose masking the misaligned regions to mitigate their impact~\cite{kong2024safnet}. Alternatively, diffusion models have been
explored to reconstruct HDR images in a generative manner~\cite{yan2023toward, hu2024generating}.
Nevertheless, these learning-based methods are greatly dependent on the training samples with HDR GTs~\cite{prabhakar2021labeled}. 
The expensive HDR GTs limit the scale of HDR datasets, which causes a bottleneck.

Therefore, the annotation-efficient methods are introduced to train a HDR reconstructing model with a small amount of HDR GTs~\cite{prabhakar2021labeled,yan2023smae}. 
The common strategy of these methods is to make use of both the data with and without HDR GTs.
Nevertheless, these methods have neither successfully repaired nor selected high-quality pseudo GTs, nor effectively learned to deghost the HDR results, ultimately resulting in artifact-containing outcomes as shown in Fig.~\ref{fig_0}.
In contrast to previous annotation-efficient methods, we achieve a better performance by masking out the unreliable parts of pseudo GTs at both patch and pixel levels.

\subsection{Semi-Supervised Learning}
Semi-supervised learning (SSL) aims to train models with limited annotations~\cite{qiao2018deep,tarvainen2017mean,sohn2020fixmatch,chen2022label,wang2024semi}. Among them, recent methods follow a pseudo-labeling paradigm~\cite{rasmus2015semi,laine2016temporal,sohn2020fixmatch,chen2022label}, which our work also follows. 
The pseudo-labeling methods train a student model with a mix of human annotations and pseudo annotations generated from a teacher model. 
These methods significantly enhance various computer vision tasks: detection~\cite{chen2022label,tang2021humble}, segmentation~\cite{fan2022ucc,wu2024instance}, and computational photography~\cite{pan2024pseudo,pan2024pseudo1}.

Nevertheless, when used for new tasks, vanilla pseudo-labeling approaches, such as the mean teacher~\cite{tarvainen2017mean}, can face challenges due to confirmation bias.
The confirmation bias means the student model could learn errors from the pseudo annotations~\cite{tarvainen2017mean}.
Current approaches aim to mitigate the impact of confirmation bias by carefully selecting pseudo labels~\cite{chen2022label,liu2022unbiased}.
Our work addresses the issue of confirmation bias
by conducting the pixel-level uncertainty estimation~\cite{he2019bounding} for pseudo HDR GTs. 
The trusted regions of the pseudo HDR GTs are selected with bi-level thresholds. 

\begin{figure*}[t]
  \includegraphics[width=\textwidth]{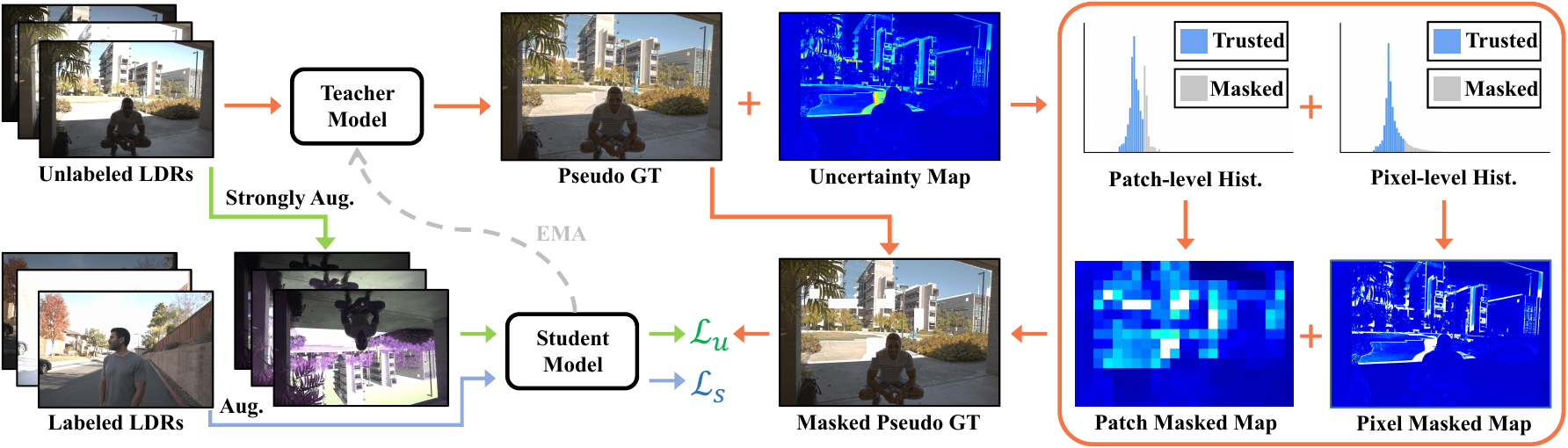}
  \centering

  \caption{\textbf{The pipeline of the proposed annotation-efficient HDR image reconstructing framework.} The framework follows the teacher-student structure. The teacher branch predicts both the pseudo HDR GTs and the corresponding uncertainty maps. The uncertainty map evaluates the pixel reliability of pseudo GTs. Thus, we can discard the uncertain regions of the pseudo HDR GTs. In this way, the student can hardly be affected by the teacher's mistakes. We mask out the uncertain regions of the pseudo GTs from both the patch- and pixel-level. The student model learns from the masked pseudo GTs with the unsupervised loss ($\mathcal{L}_u$). The student also learns from the real GTs with the supervised loss ($\mathcal{L}_s$). Then, the teacher model is updated by student model via exponential moving average (EMA)~\cite{tarvainen2017mean}.
  }
  \label{fig:pipline}

\end{figure*}

\section{Methodology}
In this section, we firstly introduce the setting of semi-supervised HDR image reconstructing. Secondly, we review the definition of multi-frame HDR image reconstructing. Then, we provide an overview of our teacher-student structure and how to exploit unlabeled data. Finally, we discuss how to select the trusted region in HDR GTs via the proposed bi-level uncertain area masking strategy.

\subsection{Semi-Supervised Setting}
\label{sec:Annotation-Efficient Data Distribution}
The proposed semi-supervised HDR image reconstructing framework uses paired data $D^l=\{(X^l, y)_i\}_{i=1}^{N^l}$ and unpaired data $D^u=\{(X^u)_i\}_{i=1}^{N^u}$ as training data, where 
$X^l$ and $X^u$ represents the LDR image burst with and without the corresponding GT HDR image $y$. $N^l$ and $N^u$ are the total number of data with and without HDR GT, respectively. 

\subsection{Multi-Frame HDR Image Reconstructing}
\label{sec:Multi-frame HDR deghosting}
Multi-frame HDR image reconstructing focuses on restoring a ghost-free HDR image from a set of differently exposed LDR images. 
Each input LDR image set $X$ has three input LDR images $\left(x_1, x_2, x_3\right)$ with varied exposures.
The middle frame $x_2$ is treated as the reference image. To project the LDR image set $X$ to the HDR domain, we use gamma correction to generate the gamma-corrected image set $\check{X}$ :
\begin{equation}
\check{x}_i=\frac{\left(x_i\right)^\gamma}{t_i}\,, \quad i=1,2,3    
\end{equation}
where $t_i$ indicates the exposure time of $x_i$, and $\gamma$ is the gamma correction parameter. 
Following \cite{kalantari2017deep}, we 
concatenate the original LDR images $X$ and the gamma-corrected images $\check{X}$ to obtain the final input $\widehat{X} = \left( X, \check{X} \right)$. Both $X^l$ and $X^u$ go through all above operations. The network $\Phi\left(\cdot\right)$ is then defined as:
\begin{equation}
\label{eq2}
r=\Phi(\widehat{X}; \theta)\,,
\end{equation}
where $r$ denotes the reconstructed HDR image, and $\theta$ is the network parameter to be optimized. For the network, GFHDR is selected as the teacher and student of our approach, detailed by \cite{liu2022ghost}. 

\subsection{Model Overview}
\label{sec:model overview}

Fig.~\ref{fig:pipline} shows the proposed framework which follows the teacher-student pseudo-labeling paradigm. This teacher-student framework contains two networks with the same structure but different weights, which are denoted as $\theta_t$ and $\theta_s$ for the teacher and student, respectively.
The teacher branch generates pseudo GTs for the data without HDR GTs. The pseudo GTs are then used as the supervised signal to train the student.
Parts of the pseudo GTs contain ghosting or noise, which may misguide the student. Therefore, we design an uncertainty branch to evaluate the quality of the reconstructed HDR images.
Then with the uncertainty map, we can mask the unreliable parts of the pseudo GTs. Different from the student being updated by gradients, the teacher is updated by the student weights via the exponential moving average (EMA)~\cite{tarvainen2017mean}:
\begin{equation}
\theta_t^i \leftarrow \alpha \theta_t^{i-1}+(1-\alpha) \theta_s^i\,,
\end{equation}
where $\alpha$ is a hyper-parameter to control the updating speed. $\theta_t^i$ and $\theta_s^i$ are the parameters of the teacher and student in the $i$-th iteration, respectively. 
The student is trained with supervised losses $\mathcal{L}_s$ on images with GTs and unsupervised losses $\mathcal{L}_u$ on images without GTs. As HDR images are normally viewed after tone-mapping $\mathcal{T}(\cdot)$,  before computing the losses, both the predicted HDR images and the HDR GTs are tone-mapped by a $\mu$-law function:
\begin{equation}
\label{tonemaping}
\mathcal{T}(I)=\frac{\log (1+\mu I)}{\log (1+\mu)}\,,
\end{equation}
where $I$ can be $r$ or $y$, $\mu$ is a hyper-parameter.
Following \cite{liu2022ghost}, we use
$\mathcal{L}^r$ loss and perceptual loss
$\mathcal{L}^v$~\cite{johnson2016perceptual} as the optimization targets.
On the data $\widehat{X^l}$ with real GTs, the supervised losses can be computed as:
\begin{equation}
\mathcal{L}_s^r=\bigr\|\mathcal{T}\bigr(\Phi(\widehat{X^l}; \theta_s)\bigl)-\mathcal{T}\bigr(y\bigl)\bigl\|_1,
\end{equation}
\begin{equation}
\mathcal{L}_s^v=\sum_j\big\|\Psi_j\big(\mathcal{T}\big(\Phi(\widehat{X^l}; \theta_s)\big)\big)-\Psi_j\big(\mathcal{T}\big({y}\big)\big)\big\|_1\,,
\end{equation}
where $\Psi\left(\cdot\right)$ denotes the activation feature maps extracted from a pre-trained VGG-16 network~\cite{simonyan2014very}, and $j$ denotes the $j$-th layer. On the data $\widehat{X^u}$ without GTs, the unsupervised losses can be calculated as:
\begin{equation}
\mathcal{L}_u^r=\big\|\mathcal{T}\big(\Phi(\widehat{X^u}; \theta_s\big)-\mathcal{T}\big({y^p}\big)\big\|_1\,,
\end{equation}
\begin{equation}
\mathcal{L}_u^v=\sum_j\big\|\Psi_j\big(\mathcal{T}\big(\Phi(\widehat{X^u}; \theta_s)\big)-\Psi_j\big(\mathcal{T}\big({y^p}\big)\big)\big\|_1\,,
\end{equation}
where $y^p = \Phi(\widehat{X^u}; \theta_t))$ is the pseudo HDR GT.
Together with the supervised uncertainty loss $\mathcal{L}_s^k$ and the unsupervised uncertainty loss $\mathcal{L}_u^k$ which will be defined in the following section, the total loss of our framework is:
\begin{equation}
\mathcal{L} = \mathcal{L}_s^r + \lambda_v\mathcal{L}_s^v + \mathcal{L}_s^k +\lambda_u\left(\mathcal{L}_u^r+ \lambda_v\mathcal{L}_u^v + \mathcal{L}_u^k\right)\,,
\end{equation}
where $\lambda_u$ and $\lambda_v$ are hyper-parameters. The $\lambda_u$ is used to trade off between the supervised and the unsupervised loss. 

\begin{figure}[t]
  \includegraphics[width=\linewidth]{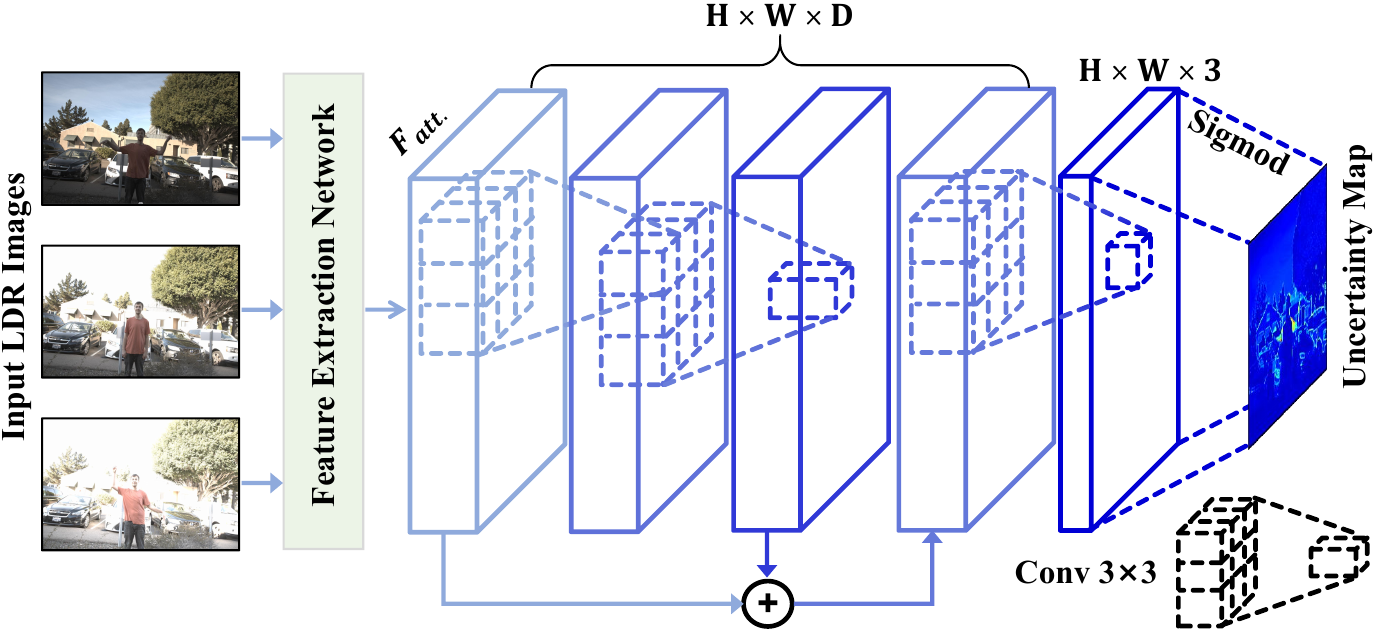}
  \centering

  \caption{\textbf{The architecture of the judge network.} The input LDR images are first fed to the attention-based feature extraction network~\cite{liu2022ghost} to obtain the fused feature $F_{att}$. Then the three $3\times3$ convolution layers with a skip addition connection generate the uncertainty map.}
  \label{fig:scorenet}

\end{figure}

\subsection{Bi-Level Uncertain Area Masking}
\label{sec:Uncertainty pseudo label filtering}

The challenge of our pseudo-labeling framework lies in how to discard artifacts of the pseudo HDR GTs. We argue to explore trusted pseudo GTs so that the student can learn from the reliable regions of the pseudo GTs. This ensures that the student is unlikely to break down
and can perform better. Based on the EMA, the better student leads to the better teacher. So, the updated teacher can produce trusted pseudo GTs on previously uncertain regions. This makes a positive feedback loop. But there is no indicator, \eg, the confidence score in classification task, to evaluate the quality of pseudo labels in this field. 
Therefore, in this paper, we introduce a judge network to directly provide corresponding evaluation scores along with the predicted HDR pixels. Considering that the HDR image reconstructing is essentially a dense regression task, we follow the idea of~\cite{he2019bounding} to treat the predictions and the regression targets as Gaussian distributions and Dirac delta functions respectively. Thus, the uncertainty-aware loss for our task can be written as:
\begin{equation}
\mathcal{L}^k = \frac{1}{n_h}\sum_i(\frac{(\bar{h}_i-h_i)^2}{2 \sigma_i^2}+\frac{1}{2} \log (\sigma_i^2))\,,
\end{equation}
where the $\bar{h}$ and $h$ are respectively predicted and real/pseudo GT HDR image pixel values, $\sigma_i$ is the uncertainty score 
of the $i$-th pixel, $n_h$ is the total number of pixels.
When the value $\bar{h}$ is estimated inaccurately, the network is expected to be able to predict larger $\sigma$ so that $\mathcal{L}^k$ will be lower. 

As shown in Fig.~\ref{fig:scorenet}, we design a judge network with three-layer $3\times3$ convolution, one skip connection, and a sigmoid function to predict uncertainty scores. It takes feature 
$F_{att}$ 
extracted by the spatial attention module in GFHDR~\cite{liu2022ghost} as input to obtain the uncertainty map of the predicted HDR image.
The uncertainty map explicitly shows the quality of the HDR image by providing the magnitude of the possible deviation of predicted values. 

Based on the uncertainty map, we 
mask out the unreliable regions of the pseudo GTs 
from the patch and pixel levels to 
ensure the reliability of the pseudo GTs 
during utilization.
Before training, LDR images and HDR images need to be cropped into patches, we reject the pseudo GT patches
that have ghosting or artifacts. In this process, we first calculate the mean value of the uncertainty map 
at the patch level, then perform global normalization to get the normalized patch-level uncertainty score $S_{pa}$. Furthermore, we 
obtain a patch-level mask $M_{pa}$ according to a pre-set threshold $\tau_{pa}$. The patch-level mask is calculated by:
\begin{equation}
M_{pa}^i= \begin{cases}1 & S_{pa}^i<\tau_{pa} \\
0 & s_{pa}^i>\tau_{pa} \\
\end{cases}\,,
\end{equation}
where $1$ means that the corresponding patch data is used as training data and vice versa.

However, there may still be small areas of ghosting and artifacts in the remaining patches, which requires more refined filtering 
at the pixel level. Therefore, we first calculate the mean value of each pixel in RGB channel and 
normalize the values in all pseudo labels to obtain a normalized single-channel uncertainty map. 
A binary uncertainty mask $M_{pi}$ is then generated by a predefined threshold $\tau_{pi}$:
\begin{equation}
M_{pi}^j= \begin{cases}1 & S_{pi}^j<\tau_{pi} \\
0 & s_{pi}^j>\tau_{pi} \\
\end{cases}\,.
\end{equation}
With the help of this uncertainty mask, we can accurately filter out low-quality pixels with 0 values in $M_{pi}$, which are not used to calculate losses during pseudo label training.

\begin{table*}

  \setlength\tabcolsep{1mm}
  \centering
  \begin{tabular}{c|c|cccccccc}
    \toprule
    Dataset & Metric & AHDRNet & ADNet & GFHDR & SCTNet & SAFNet & FSHDR & SMAE & Ours \\
    \midrule
    \multirow{2}{*}{Kalantari} 
    & PSNR-$\mu$ & 41.05±0.32 & 40.93±0.38 & \underline{41.63±0.23} & 41.58±0.27 & 37.81±0.44 &  41.40±0.13 & 41.61±0.08 & \textbf{44.04±0.07} \\
    & PSNR-$l$ & 40.61±0.10 & 40.78±0.15 &  39.85±0.42 & 39.93±0.40 & 37.95±0.65 & 41.39±0.12 & \underline{41.54±0.10} &\textbf{41.67±0.06} \\
    \specialrule{0em}{1pt}{1pt}
    \hline
    \specialrule{0em}{1pt}{1pt}
    \multirow{2}{*}{Hu} 
    & PSNR-$\mu$ & 43.42±0.44 & 43.79±0.48 &  42.97±0.17 & 43.18±0.62 & 41.47±0.37 & 43.98±0.27 & \underline{44.24±0.17} & \textbf{45.10±0.25} \\
    & PSNR-$l$ &  46.37±0.76 & 46.88±0.81 &  45.80±0.93 & 45.09±0.79 & 44.39±0.94 & 47.13±0.13 & \underline{47.41±0.12} & \textbf{47.93±0.14}\\
  \bottomrule
  \end{tabular}
  \caption{Results on Kalantari’s~\cite{kalantari2017deep} and Hu's~\cite{hu2020sensor} dataset under the semi-supervised setting of SMAE~\cite{yan2023smae}. The best and the second-best results are respectively highlighted in \textbf{bold} and \underline{underline}.}
  \label{tab:table1}

\end{table*}

\subsection{Data Augmentation} Suggested by FixMatch~\cite{sohn2020simple}, data augmentation with appropriate differences can lead to a success in SSL by incorporating
consistency learning and pseudo-labeling. 
Therefore, for the `` Strongly Aug.'' in Fig.~\ref{fig:pipline},  we propose to superimpose RGB random shuffle on top of horizontal flip and 90-degree counterclockwise rotation to generate strongly augmented unlabeled LDRs
with pseudo GTs for the unsupervised training of student. A contrast experiment on the composition form of the `` Strongly Aug.'' is provided in the 
Supplementary Materials.
In addition, to make better use of the data with HDR GTs, we follow GFHDR~\cite{liu2022ghost} 
to perform vertical flip and 90-degree clockwise rotation for augmentation.


\subsection{Details} 
The proposed semi-supervised HDR deghosting framework is trained with two stages: the warm-up stage and the semi-supervised stage. In the warm-up stage, the student only learns from the labeled data $D^l$.
During the semi-supervised stage, the teacher first initializes weights $\theta_t$ 
via copying the student's weights $\theta_s$ and is then updated by EMA along with the training step of the student. 
Compared 
with the teacher $\theta_t$ updated after every step, the pseudo HDR GTs are generated after every epoch. Then the trusted regions in pseudo HDR GTs are selected with the guidance of uncertainty maps. Finally, the student can learn from both the annotated data and trusted pseudo-labeled data. 

\begin{figure*}[!t]
  \includegraphics[width=\linewidth]{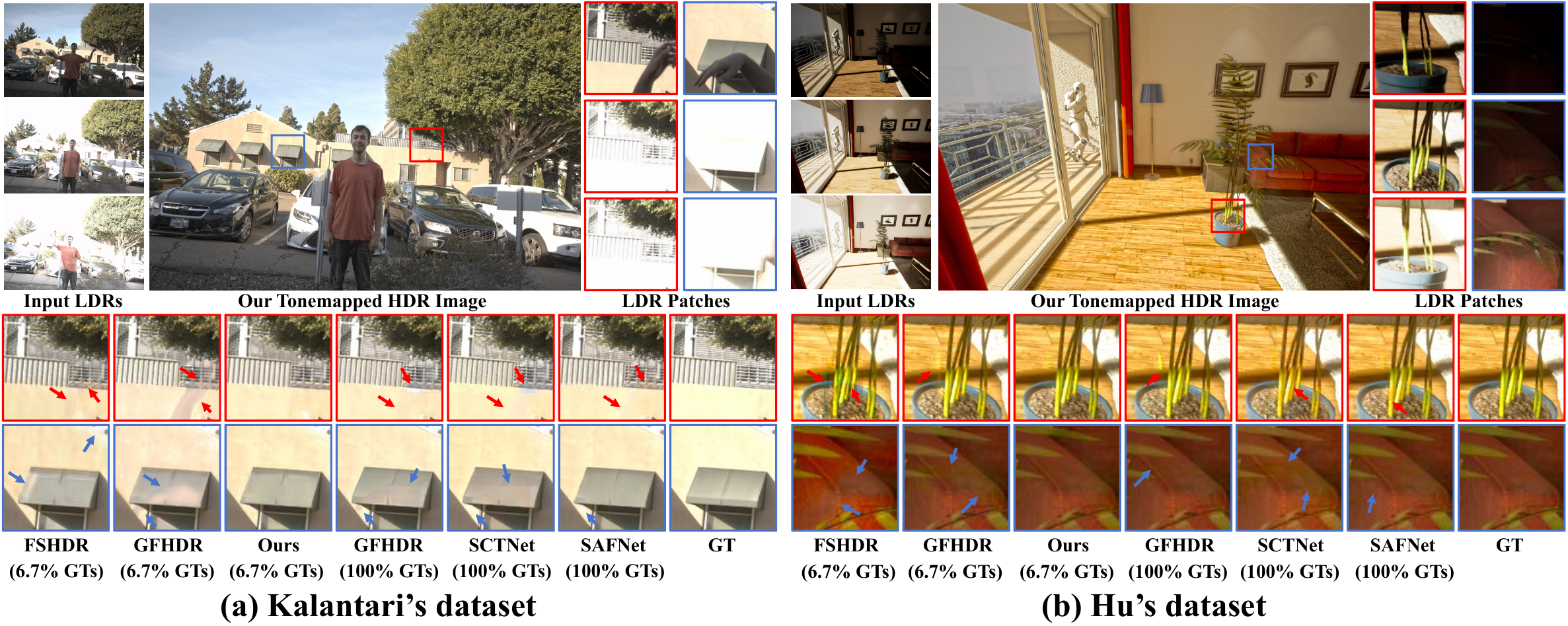}
  \centering

  \caption{\textbf{Examples of qualitative results on dataset with GT.} Subfigure
(a) and (b) present the examples of Kalantari’s~\cite{kalantari2017deep} and Hu’s~\cite{hu2013hdr} datasets, respectively.}
  \label{fig:view}
\end{figure*}

\section{Experiments}
\subsection{Dataset and Settings}
\textbf{Datasets.} Following previous methods~\cite{liu2022ghost,yan2023smae}, we use the training and testing datasets provided by Kalantari~\cite{kalantari2017deep} and Hu~\cite{hu2020sensor}.
Each of these scenes include three differently exposed input LDR images (exposure values of $\left\{-2, 0, +2\right\}$ or $\left\{-3, 0, +3\right\}$ for Kalantari's and $\left\{-2, 0, +2\right\}$ for Hu's ) which contain dynamic elements (\eg, camera motion, non-rigid movements) and a GT HDR image aligned with the medium frame captured via static exposure fusion. Moreover, we quantitatively and qualitatively evaluate our method on these two testing sets. 

\textbf{Evaluation Metrics.} 
We mainly include two kinds of objective metrics for evaluation: PSNR and SSIM. In general, we calculate PSNR-$l$, PSNR-$\mu$, SSIM-$l$ and SSIM-$\mu$ to make a quantitative score between the predicted HDR images and corresponding GTs. The $-l$ means the metric is computed directly in the linear domain, While the $-\mu$ means the HDR images are first tonemapped by Eq.~\ref{tonemaping}. 
The HDR-VDP-2~\cite{mantiuk2011hdr} is also calculated, which evaluates the HDR image pairs in both visibility and quality.

\textbf{Experimental Settings.} The proposed framework is implemented by PyTorch. During training, the input images are randomly cropped with size of $64\times64$  and stride of $32$. We use the Adam optimizer for the student, and set the initial learning rate to $2e-4$, $\beta_1$ to 0.9, $\beta_2$ to 0.999, and $\epsilon$ to $1e-8$. We set $\gamma$ to 2.2, $\mu$ to 5000 for the tonemapping, $\alpha$ to 0.999 for the EMA, and $\lambda_u$ to 1 for the unsupervised loss. The $\tau_{pa}$ and $\tau_{pi}$ are both set to 0.4 for mask threshold. Sensitivity analyses about the selection of $\alpha$, $\tau_{pa}$ and $\tau_{pi}$ are in the 
Supplementary Materials. 
The framework is trained for 200 epochs: 30 for warm-up and 170 for semi-supervised training. With $N^l = 5$ and a batch size of 64, training takes about two days on two NVIDIA 3090Ti GPUs.
Since the GFHDR is selected as our teacher and student model, our model's inference time is the same as GFHDR, which can be referred to \cite{liu2022ghost} for more details.

\begin{table}[!t]

  \setlength\tabcolsep{1pt} 
  \centering
  \begin{tabular}{c|c|ccccc}
    \toprule
    Setting & Methods & PSNR-$\mu$ & PSNR-$l$ & SSIM-$\mu$ & SSIM-$l$ & HV2 \\
    \midrule
    \specialrule{0em}{1pt}{1pt} 
    100\%&  Kalantari &  42.67 &  41.23 &  0.9888 &  0.9846 &  65.05 \\
    GTs&  DeepHDR &  41.65 & 40.88 &  0.9860 &   0.9858 &  64.90 \\
    &  AHDRNet &  43.63 & 41.14 &  0.9900 &  0.9702 &  64.61 \\
    & FSHDR & 41.92 & 41.79 & 0.9851 & 0.9876 &  65.30 \\
    &  HDR-GAN & 43.92 & 41.57 & 0.9905 & 0.9865 &  65.45 \\
    & SMAE & 41.97 & 41.68 & 0.9895 & 0.9889 & - \\
    & GFHDR & 44.32 & 42.18 & 0.9916 & 0.9884 & 66.03 \\ 
    &HyHDRNet&44.64&42.47&0.9915&0.9894& 66.05\\
    &SCTNet& 44.49&42.29&\underline{0.9924}&0.9887& \underline{66.65}\\
    &DHRNet& \textbf{44.76}&\underline{42.59}&0.9919&\underline{0.9906}& 66.54\\
    &SAFNet& \underline{44.66}& \textbf{43.18} & \textbf{0.9932} & \textbf{0.9917} & \textbf{66.93} \\
    \specialrule{0em}{1pt}{1pt}
    \hline
    \specialrule{0em}{1pt}{1pt} 
     6.7\%& FSHDR & 41.94 & 40.80 & 0.9847  & 0.9860 & \underline{64.45} \\
     GTs& SMAE & 41.61  & \underline{41.54} & 0.9880  & \underline{0.9879} & - \\
    & GFHDR & \underline{42.72} & 39.80 & \underline{0.9907} & 0.9862 & 63.74  \\
    & Ours & \textbf{44.15} & \textbf{42.55} & \textbf{0.9920} & \textbf{0.9890} & \textbf{64.87} \\
    \specialrule{0em}{1pt}{1pt}
    \hline
    \specialrule{0em}{1pt}{1pt}
    13.5\%& FSHDR & 42.51 & \underline{40.97} & 0.9868 & 0.9855 & 64.25  \\
    GTs& GFHDR & \underline{43.32} & 40.80 & \underline{0.9908}  & \underline{0.9867} & \underline{64.91} \\
    & Ours & \textbf{44.16}  & {\textbf{42.81}} & \textbf{0.9920}  & \textbf{0.9899} & \textbf{65.89} \\
    \specialrule{0em}{1pt}{1pt}
    \hline
    \specialrule{0em}{1pt}{1pt}
     27.0\%& FSHDR &  43.03 & \underline{41.97} & 0.9891 & 0.9863 & \underline{65.53} \\
    GTs& GFHDR & \underline{43.14} & 41.09 & \underline{0.9911} & \underline{0.9880} & 65.17 \\
    & Ours & \textbf{44.21}  & \textbf{42.86} & {\textbf{0.9921}}  & {\textbf{0.9905}} & \textbf{66.10} \\
    \specialrule{0em}{1pt}{1pt}
    
  \bottomrule
  \end{tabular}
  \caption{Further semi-supervised and supervised results on Kalantari's
  ~\cite{kalantari2017deep}
  dataset. The best and the second-best results under different settings are respectively highlighted in \textbf{bold} and \underline{underline}. 
  HV2 represents HDR-VDP-2.}
  \label{tab:table2}
\end{table}

\subsection{Results on Kalantari's and Hu's Dataset} Table~\ref{tab:table1} presents a comparison between our method and several state-of-the-art DNN-based approaches under the semi-supervised setting introduced by SMAE~\cite{yan2023smae}. 
 For a fair comparison, all results are reported with means and 95 $\%$ margin of variations across 5 runs. As shown in Table~\ref{tab:table1}, our method significantly outperforms the baseline method and attains the SOTA performance on all metrics of two datasets. 
 The proposed method achieves superior performance over the second-best approach, with improvements of 2.41 dB and 0.13 dB in terms of PSNR-$\mu$ and PSNR-$l$ on the Kalantari's dataset, and 0.86 dB and 0.52 dB on the Hu's dataset, respectively.

\subsection{More Quantitative Results on Kalantari's Dataset}
The additional experiments on the Kalantari's dataset in Table~\ref{tab:table2} are conducted under different semi-supervised settings.
Three different semi-supervised settings are chosen depending on the amount of annotated data used: $N^l = 5$ (6.7\% GTs), $N^l = 10$ (13.5\% GTs), $N^l = 20$ (27\% GTs). All the remaining annotated data is treated as unannotated data.
 The results of supervised SOTA methods trained with total 74 (100\% GTs) annotated data are also presented.
 As shown in Table \ref{tab:table2}, our method consistently achieves the SOTA performance across all semi-supervised settings. Moreover, our method achieves performance comparable to the fully supervised SOTA methods using only 6.7\% labeled data. What's more, with a total of 27\% labeled data, our method achieves the global second-best results in PSNR-$l$ metric.
 Briefly, all of the above comparison results further confirm the effectiveness of our method.


\subsection{Qualitative results}
As shown in Fig.~\ref{fig:view}(a) and (b), we conduct qualitative comparisons on Kalantari's dataset and Hu's dataset. In fact, due to the presence of complex factors such as saturation and large motion, achieving color-accurate and ghost-free reconstruction of HDR images is challenging. When the labeled GTs are limited, FSHDR uses reverse optical flow to generate additional pseudo-labeled synthetic data pairs for training. However, the domain gap between the real and synthesized data may misguide the model and lead to unnatural new artifacts and color distortions. Baseline method GFHDR struggles to learn effective alignment from limited GT
data, resulting in severe ghosting artifacts. Thanks to the more trusted pseudo GT data and the strong data augmentation-driven consistency learning, our method can learn effective information from unlabeled data,
achieving performance close to supervised methods.

\begin{figure*}[!t]

  \includegraphics[width=\linewidth]{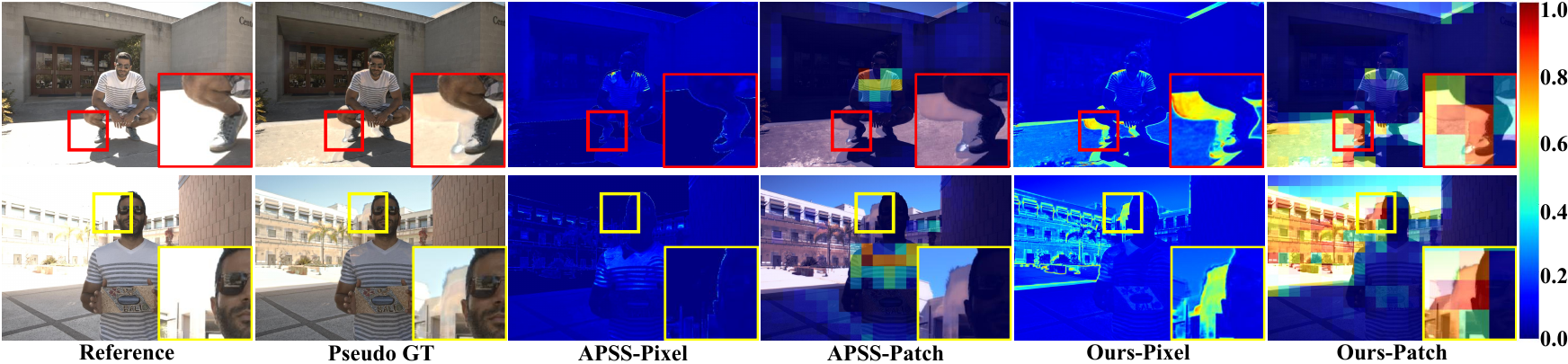}
  \centering

  \caption{
  \textbf{Qualitative comparison of pseudo HDR GT masking strategies.}
  Boxed and zoomed regions highlight notable artifacts in pseudo GTs. Columns 3–6 show APSS uncertainty maps in SMAE and our method at pixel and patch levels. To enhance visual perception, the heat maps are normalized, higher colorbar values indicate more artifacts in pseudo GTs.
  }
  \label{fig:select}

\end{figure*}

\subsection{Analyses}
\label{subsec:Analyses}
\textbf{Ablation studies.} 
As shown in Table~\ref{tab:ablation}, we conduct ablation studies on the modules used in our method on Kalantari's dataset with 6.7\% GTs training data to confirm the roles of each module. Among the modules, the GFHDR~\cite{liu2022ghost} is selected as our baseline (BL) module, and the performance is presented in line \romannumeral1. By comparing the convergence performance with different modules, we can observe:

\textit{(a) The vanilla MT does not work well in our task.}
Compared with line \romannumeral1, barely incorporating a vanilla mean teacher (MT) network in line \romannumeral2~without rectifying pseudo GTs does not work. The marginal performance improvement indicates that the model with the only vanilla MT struggles to learn useful information from unlabeled data, which is also known as suffering from confirmation bias.
Accordingly, refining the pseudo labels is vital in our semi-supervised framework.

\textit{(b) The uncertainty loss $\mathcal{L}_u^k$  provides uncertainty scores only.}
The loss $\mathcal{L}_u^k$ can supervise the learning of the network prediction results and the corresponding scores. The comparison between lines \romannumeral3~and \romannumeral2~shows that the loss function does not affect the results positively or negatively. However, as depicted in the last two columns of Fig.~\ref{fig:select},  the corresponding scores are accurately estimated, which provides a foundation for us to select trusted pseudo GTs.

\textit{(c) The filtration at two levels is better.}
With the corresponding pixel-level scores, the pseudo GTs can be filtered to remove regions with artifacts at two levels, namely the pixel level and patch level. In the comparison between the lines \romannumeral4~to \romannumeral6~with line \romannumeral2, the better convergence performance in line \romannumeral6~indicates that simultaneous masking at both the pixel and patch levels allows for a more precise filtration of pseudo GTs, which can be the better rectification strategy. 


\textit{(d) The strong augmentations for pseudo-labeled data training are necessary.}
Following~\cite{sohn2020fixmatch}, pseudo-labeled data learning in the student needs to be driven by strong augmentations. With the proposed strong augmentations, our method in line \romannumeral7~composed of all modules, achieves a significant improvement over the baseline. 

\begin{table}[!t]
  \setlength\tabcolsep{1mm}
  \centering
  \begin{tabular}{c|cccccc|cc}
    \toprule
    No. & BL & MT & $\mathcal{L}^k$ & PaM & PiM & SAug & PSNR-$\mu$ & SSIM-$\mu$\\
    \midrule
    \romannumeral1 & $\checkmark$ & & & & & & 42.7220 & 0.9907 \\
    \romannumeral2 & $\checkmark$ & $\checkmark$ & & & & & 42.6509 & 0.9909 \\
    \romannumeral3 & $\checkmark$ & $\checkmark$ & $\checkmark$ & & & & 42.8893 & 0.9905 \\
    \romannumeral4 & $\checkmark$ & $\checkmark$ & $\checkmark$ & $\checkmark$ & & & 43.3528 & 0.9908 \\
    \romannumeral5 & $\checkmark$ & $\checkmark$ & $\checkmark$ & & $\checkmark$ & & 43.2513 & 0.9909 \\
    \romannumeral6 & $\checkmark$ & $\checkmark$ & $\checkmark$ & $\checkmark$ & $\checkmark$ & & 43.4787 & 0.9909 \\
    \romannumeral7 & $\checkmark$ & $\checkmark$ & $\checkmark$ & $\checkmark$ & $\checkmark$ &  $\checkmark$ & \textbf{44.0378} & \textbf{0.9915} \\
    \bottomrule
  \end{tabular}

  \caption{Ablation studies. PaM, PiM and SAug represent patch-level masking, pixel-level masking, and strong augmentations.}
  \label{tab:ablation}

\end{table}

\textbf{The difference in pseudo GTs scoring strategy.}
When the teacher model's capacity is insufficient, the generated pseudo GTs often exhibit ghosting artifacts as highlighted in the second column of the Fig.~\ref{fig:select}. The SMAE utilizes the similarity between the generated HDR image and the reference frame as the scoring criterion, while ignoring saturated regions. This results in low scores that overlook ghosting artifacts and hinder the learning of well-reconstructed regions, which explains the inferior performance of APSS in line \romannumeral1~of Table 4.
We propose the judge network 
and employ the $\mathcal{L}^k$ loss to guide the learning of uncertainty scores, enabling the model to assess its own predictions in an uncertainty-aware manner. This facilitates the evaluation of pseudo HDR ground truths 
and effectively identifies ghosting-affected areas, as illustrated in the fifth and sixth columns of Fig.~\ref{fig:select}. 
Consequently, the model’s performance can be effectively improved by leveraging our proposed dual-threshold filtering strategy to select reliable pseudo GTs.
\begin{table}[!t]

  \setlength\tabcolsep{1mm}
  \centering
  \begin{tabular}{c|ccccc|cc}
    \toprule
    No. & BL & MT & FSHDR & APSS & Ours & PSNR-$\mu$ & SSIM-$\mu$\\
    \midrule
    \romannumeral1 & $\checkmark$ & $\checkmark$ & & $\checkmark$ & & 42.7956 & 0.9910 \\
    \romannumeral2 & $\checkmark$ & & $\checkmark$ & & & 43.5640 & 0.9914 \\
    \romannumeral3 & $\checkmark$ & $\checkmark$ & & & $\checkmark$ & \textbf{44.0378} & \textbf{0.9915} \\
    \bottomrule
  \end{tabular}

  \caption{Ablation studies. The scoring strategy of the APSS in SMAE  and semi-supervised strategy in FSHDR are conducted with the same settings as ours.}
  \label{tab:strategy}

\end{table}

\textbf{The strategies in semi-supervised learning.}
In lines \romannumeral2~and \romannumeral3~of Table~\ref{tab:strategy}, we present the performance of the BL module trained in FSHDR or ours strategy. FSHDR utilizes reverse optical flow to generate additional synthetic pseudo-labeled data, which, however, disrupts the original distribution of real data. In contrast, our method does not change the original distribution and achieves superior performance. 

\section{Conclusion}
Paired LDR-HDR data is rare, which limits the development of learning-based HDR image reconstructing. Previous works have attempted to solve this problem from the perspective of semi-supervised learning but struggle to obtain good enough pseudo HDR GTs. To deal with this limitation, we develop a novel teacher-student framework where the student learns from the EMA version of itself, \ie, the teacher model, and propose a bi-level uncertain area masking strategy. In this way, the student model can learn from trusted pseudo labels, which consequently yields a better teacher model. Our proposed method achieves significantly superior performance to the state-of-the-art semi-supervised methods. Moreover, using the data with only $6.7\%$ total available GTs, we achieve comparable performance with state-of-the-art fully supervised methods.

\bibliography{aaai2026}

\end{document}